\documentclass[runningheads]{llncs}
\usepackage{graphicx}
\usepackage{amsmath}
\usepackage{orcidlink}
\usepackage{listings}
\usepackage{float} 
\usepackage{caption}
\usepackage{subcaption}
\usepackage{caption}
\usepackage{url}
\usepackage{lscape}
\usepackage{longtable}
\usepackage{cite}

\begin{document}
\title{Estimation of total body fat using symbolic regression and evolutionary algorithms.}
\titlerunning{TBFAT by SR}

\author{Jose-Manuel Muñoz\inst{1}\orcidID{0009-0007-7134-8708} \and Odin Morón-García \orcidID{0000-0002-5275-4520} \and  J. Ignacio Hidalgo\inst{2}\orcidID{0000-0002-3046-6368} \and Omar Costilla-Reyes \inst{3}\orcidID{ 0000-0001-8331-7262 }}

 \authorrunning{al et al.}

 \institute{Instituto tecnológico de Tijuana \and Universidad Complutense de Madrid, Spain.
\email{hidalgo@ucm.es} \and MIT} 

\maketitle              

\begin{abstract}
Body fat percentage is an increasingly popular alternative to Body Mass Index to measure overweight and obesity, offering a more accurate representation of body composition. In this work, we evaluate three evolutionary computation techniques, Grammatical Evolution, Context-Free Grammar Genetic Programming, and Dynamic Structured Grammatical Evolution, to derive an interpretable mathematical expression to estimate the percentage of body fat that are also accurate. Our primary objective is to obtain a model that balances accuracy with explainability, making it useful for clinical and health applications. We compare the performance of the three variants on a public anthropometric dataset and compare the results obtained with the QLattice framework. Experimental results show that grammatical evolution techniques can obtain competitive results in performance and interpretability.

\end{abstract}

\section{Introduction}
Body Mass Index (BMI) has traditionally served as the primary indicator of obesity and related health conditions. BMI is easy to compute and use. However, it lacks precision in differentiating between fat and lean mass, leading to limitations in its predictive accuracy for individual health outcomes. This limitation has produced interest in alternative metrics, such as body fat percentage (BFP), which offers a fine assessment of body composition. BFP allows measuring metabolic health, cardiovascular risk, and other obesity-related complications.

Several methods have been developed to estimate BFP, ranging from laboratory-based techniques to predictive equations. Usual methods include dual-energy X-ray absorptiometry (DXA) \cite{pietrobelli1996dual}, bioelectrical impedance analysis (BIA) \cite{johnson2016bioelectrical}, and underwater weighing \cite{nickerson2017comparison}, providing acceptable accuracies but requiring specialized equipment and controlled environments. For a more accessible and inexpensive assessment, researchers have proposed various predictive models based on anthropometric measurements \cite{swainson2017prediction}. These models rely on combinations of waist circumference, other anthropometric features, and demographic variables such as age and gender. However, deriving models that are both clinical interpretable and accurate expressions remains a challenge.

Recently, the QLattice framework \cite{brolos2021approach} has been used to obtain a data-driven approach to derive mathematical expressions for estimating BFP, yielding promising initial results but leaving room for improvement in both accuracy and interpretability \cite{schnur2023information}. The study utilizes publicly available anthropometric data from the National Health and Nutrition Examination Survey (NHANES) of the US Centers for Disease Control and Prevention (CDC), a comprehensive dataset that represents a wide demographic cross-section of the population\cite{stierman2021national} to study nutrition and its impact in metabolic health.

In this paper, we investigate three evolutionary computation methods guided by grammars: Grammatical Evolution (GE)\cite{Ryan1998}, Context-Free Grammar Genetic Programming (CFG-GP) \cite{whigham1995grammatically}, and Dynamic Structured Grammatical Evolution (DSGE) \cite{lourencco2019structured}. They allow for helpful configuration of the solution space and generate symbolic expressions that can potentially enhance both predictive performance and model transparency. The main contributions of this work are:
\begin{itemize}
\item We systematically evaluate GE, CFG-GP, and DSGE in the derivation of interpretable models for estimating body fat percentage using different grammars. 
\item We explore the convergence and performance of different configurations based on the maximum depth tree.
\item  We analyze interpretability of the solutions.
\end{itemize}

The rest of this paper is structured as follows: Section \ref{sec:related_work} presents research the estimation of BFP. Section \ref{sec:BFP_Estimation_by_Grammars} describes the grammar-based evolutionary computation techniques of this study. Section \ref{sec:exp_results} shows the performance results and convergence analysis for the different algorithms, and Section \ref{sec:conclusions} concludes the paper.

\section{Related Work}
\label{sec:related_work}

Estimating BFP is essential for reckoning health risks associated with obesity and related conditions. A range of methods has been developed, from laboratory techniques to more accessible predictive models that use anthropometric data. Lab techniques involve advanced imaging or physical measurement techniques. DXA is one of the most accurate methods. It works by passing low-dose X-rays through the body, DXA differentiates between bone, lean mass, and fat mass, providing precise measurements of body composition \cite{pietrobelli1996dual}. However, DXA is costly, requires specialized equipment, and exposes participants to low levels of radiation, limiting its use to controlled settings.
   
Bioelectrical Impedance Analysis (BIA) estimates BFP by measuring the resistance of body tissues to a small electrical current. It provides an indirect a less precise estimation of the body composition. 
A practical alternative for estimating BFP is the use of predictive or estimation models that use anthropometric data. These models rely on easily measurable variables like height, weight, waist circumference, and sometimes skinfold thickness. Several methods have been applied for modelling. Regression-based equation discovery methods, include the Jackson-Pollock skinfold equationand the Deurenberg formulas, which uses BMI, age, and gender to estimate body fat \cite{deurenberg2002validation}. While straightforward, these equations may lack accuracy across diverse populations and body types.

The work of Schnur et al. \cite{schnur2023information} explores the use of SR through the QLattice framework to develop interpretable models for BFP. This approach focuses on creating models that can easily be understood, facilitating their application in clinical contexts where interpretability is key for decision-making. In our work, we extend and complement the work of Schnur et al., by exploring different variants of genetic programming, such as GE, CFG-GP and DSGE to obtain interpretable solutions for BFP estimation.

Our goal is to replicate the fitness results achieved in previous studies, but also to evaluate the interpretability of the models generated through evolutionary techniques. While QLattice offers a framework for SR that prioritizes clarity in the models, the evolutionary techniques explored in this study seek to find a balance between accuracy and understandability. In this way, we aim not only to verify whether these methods can match or surpass the performance of the QLattice framework, but also to explore the ability of these evolutionary methods to generate transparent and accurate models to support metabolic health assessment.

\section{Body fat estimation by grammar based GP}
\label{sec:BFP_Estimation_by_Grammars}

This study is motivated by the research of Schnur et al. \cite{schnur2023information}, which proposed a symbolic regression heuristic for estimating total body fat percentage from individual morphometric variables. Our objective is to evaluate the ability of three variants of Grammatical Guided Genetic Programming (GGGP) to generate interpretable and accurate symbolic regression models.
To this end, like Schnur et al. we used the NHANES 2017-18 dataset to train and test symbolic regression models generated with three variants of GGGP: GE, CFG-GP and DSGE. These evolutionary methods have been chosen because of their potential to find high-quality approximate solutions.

The three variants of Grammatical Evolution are implemented using PMT, a specialized Java application developed by the Absys Research group at Universidad Complutense de Madrid \cite{hidalgo2018identification}. This software framework supports the design, testing, and analysis of evolutionary algorithms, allowing researchers to experiment with different configurations and evaluate the performance of each variant effectively. 

CFG-GP is a variant of genetic programming that uses context-free grammar to guide the generation of valid solutions. This approach constrains the solutions to the structure and rules defined by grammar, ensuring that every individual in the population meets the specific syntactic constraints, which can be highly beneficial for symbolic regression, automated programming, and other applications requiring interpretable models.

GE is a type of evolutionary computation that combines principles from genetic programming with context-free grammar, allowing it to evolve computer programs, symbolic expressions, and other complex structures in a flexible and adaptable way. Instead of directly modifying program code, GE represents solutions using sequences of numbers (genomes) that map to rules in a grammar. This indirect encoding allows for greater adaptability and interpretability, especially for problems requiring domain-specific constraints.
GE has two common limitations: the locality and redundancy of its representations. High locality ensures that small changes in the genotype correspond to small changes in the phenotype, enabling effective exploration of the search space. Conversely, low locality can lead to near-random search behavior. Redundancy occurs when multiple genotypes produce the same phenotype. 

To address these challenges, Lourenço et al. \cite{lourencco2019structured} introduced Structured Grammatical Evolution (SGE). SGE employs a one-to-one mapping between genotype and grammar non-terminals. Each gene corresponds directly to a non-terminal and contains a list of integers specifying expansion options. This ensures that altering one genotypic position does not affect other non-terminals' derivations. On the other hand, SGE requires pre-processing of the input grammar to limit infinite recursion.
The following example tries to help to understand better how SGE decodes a solution with the grammar in Fig.\ref{fig:grammarbase}. In this case, the set of non-terminals would be :

\{\textless{}start\textgreater{}, \textless{}expr\textgreater{},
\textless{}op\textgreater{},\textless{}var\textgreater{},\textless{}const\textgreater{}\}, \\
so the genotype will be composed of 5 genes, one for each non-terminal (Table \ref{tab:NonTerminals-Genotype}). 
\begin{table}[!btp]
\centering
\begin{tabular}{ccccccc}
\cline{3-7}
Non-terminals & \multicolumn{1}{c|}{} & \multicolumn{1}{c|}{\textless{}start\textgreater{}} & \multicolumn{1}{c|}{\textless{}expr\textgreater{}} & \multicolumn{1}{c|}{\textless{}op\textgreater{}} & \multicolumn{1}{c|}{\textless{}var\textgreater{}} & \multicolumn{1}{c|}{\textless{}const\textgreater{}} \\ \cline{3-7} 
Genotype      & \multicolumn{1}{c|}{} & \multicolumn{1}{c|}{{[}0{]}}                        & \multicolumn{1}{c|}{{[}0, 2, 3{]}}                 & \multicolumn{1}{c|}{{[}1{]}}                     & \multicolumn{1}{c|}{{[}2{]}}                      & \multicolumn{1}{c|}{{[}0{]}}                        \\ \cline{3-7} 
              &                       &                                                     &                                                    &                                                  &                                                   &                                                    
\end{tabular}
\caption{Non-terminal and genotype}
\label{tab:NonTerminals-Genotype}
\end{table}
We present a grammar with four possible operators (\textless{}op\textgreater{}), three variables (\textless{}var\textgreater{}), and three constants (\textless{}const\textgreater{}).

The non-terminals are resolved sequentially, starting from \textless{}start\textgreater{}, to produce the phenotype. The process follows these steps:
\begin{enumerate}
    \item Select the first integer associated with the current non-terminal.
    \item Use the integer to choose the corresponding rule for that non-terminal in the grammar.
    \item Replace the non-terminal with the chosen rule and consume the integer.
\end{enumerate}
This process continues until the phenotype is fully derived. By ensuring a one-to-one correspondence between genes and non-terminals, we eliminate the need for modulo operations in the genotype-to-phenotype mapping, thereby reducing redundancy. Table~\ref{tab:GenotypeToPheno} illustrates the step-by-step derivation of a phenotype. For a more detailed explanation, we refer the interested reader to the original SGE paper~\cite{lourencco2019structured}.

\begin{table}[ht]
\centering
\begin{tabular}{|l|l|}
\hline
\textbf{Derivation Step}                                      & \textbf{Integers Remaining}                              \\ \hline
\textless{}start\textgreater{}                                & \{[0], [0, 2, 3], [1], [2], [0]\}                        \\ \hline
\textless{}expr\textgreater{}                                 & \{[], [0, 2, 3], [1], [2], [0]\}                         \\ \hline
\textless{}expr\textgreater \textless{}op\textgreater \textless{}expr\textgreater{} & \{[], [2, 3], [1], [2], [0]\}                            \\ \hline
\textless{}var\textgreater \textless{}op\textgreater \textless{}expr\textgreater{} & \{[], [3], [1], [2], [0]\}                               \\ \hline
$x_2$ \textless{}op\textgreater \textless{}expr\textgreater{} & \{[], [3], [1], [], [0]\}                                \\ \hline
$x_2$ - \textless{}expr\textgreater{}                         & \{[], [3], [], [], [0]\}                                 \\ \hline
$x_2$ - \textless{}const\textgreater{}                        & \{[], [], [], [], [0]\}                                  \\ \hline
$x_2$ - 1.0                                                  & \{[], [], [], [], []\}                                   \\ \hline
\end{tabular}
\caption{Step-by-step derivation of a phenotype.}
\label{tab:GenotypeToPheno}
\end{table}

\subsection{Methodology}\label{methodology}

\begin{figure}[ht]
	\centering
        \includegraphics[width=0.5\textwidth]{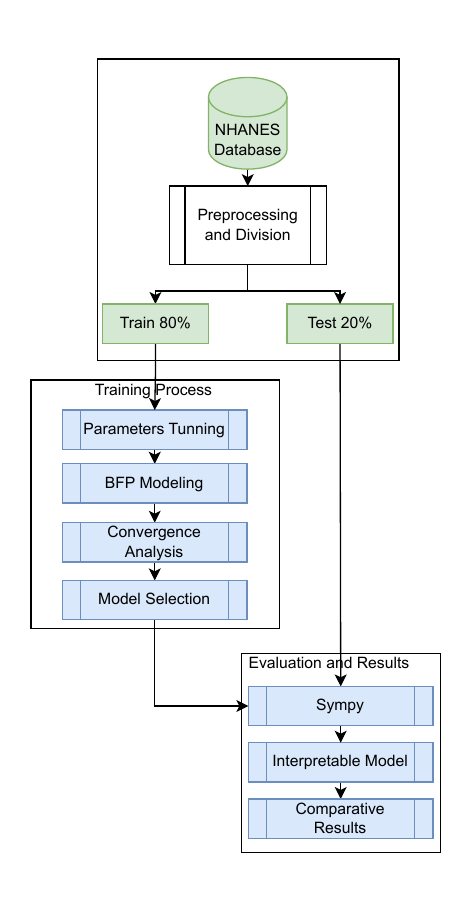}
 	\caption{Methodology used for the evaluation of  GGGP variants.}
	\label{fig:method}
\end{figure}

In the present study, a  GGGP-based methodology was implemented using the PMT software for the generation of symbolic expressions, with the purpose of modeling the underlying relationships in the NHANES dataset. The experimentation was developed following the protocol illustrated in Figure \ref{fig:method}, which included the initial segmentation of the dataset into 80\%  and 20\% for training and testing, respectively, followed by a systematic process of parameter learning in the three  GGGP variants incorporated in PMT. This optimization process was executed with the specific objective of being competitive with the performance previously achieved using the Qlattice framework.
The research was structured in multiple interconnected stages, starting with the construction of symbolic expressions for the modeling of BFP using the three GGGP variants, continuing with a convergence analysis to identify the most robust and consistent models. The methodology culminated in an evaluation on the validation set, where the selected models were tested on unseen data and subsequently translated into an interpretable form using SymPy software. The final models obtained in each GGGP variant were compared based on their fitness in R² on the test data, contrasting them with the best model generated by the Qlattice framework in the previous work by Schnur et al \cite{schnur2023information}, which allowed validating the effectiveness of the proposed approach.

\label{sec:methodology}

\begin{figure}[ht]
    \centering
\begin{lstlisting}[mathescape]
<start> ::= <expr> #(0)

<expr> ::= <expr> <op> <expr> #(0)
        | (<expr> <op> <expr>) #(1)
        | <var> #(2)
        | <const> #(3)

<op> ::= + #(0)
       | - #(1)		
       | * #(2)
       | \eb_div_\eb #(3)

<var> ::= $x_0$ #(0)
       | $x_1$ #(1)
       | $x_2$ #(2)

<const> ::= 1.0 #(0)
         | 0.1 #(1)
         | 10 #(2)
\end{lstlisting}  
\caption{Base Grammar used in this work.}
\label{fig:grammarbase}
\end{figure}

\begin{figure}[ht]
    \centering
\begin{lstlisting}[mathescape]
<start> ::= <expr> #(0)

<expr> ::= <expr> <op> <expr> #(0)
        | (<expr> <op> <expr>) #(1)
        | (<expr> <op> <expr>) #(2)
        | (<expr> <op> <var>) #(3)
        | (<var> <op> <var>) #(4)
        | (<cte> <op> <expr>) #(5)
        | (<cte> <op> <var>) #(6)
        | <var> #(7)
        | <var> #(8)
        | <var> #(9)
        | (<cte> <op> <var>) #(10)
        | (<cte> <op> <var>) #(11)
        | (<cte> <op> <var>) #(12)

<op> ::= + #(0)
       | - #(1)		
       | * #(2)
       | \eb_div_\eb #(3)

<var> ::= $x_0$ #(0)
       | $x_1$ #(1)
       | $x_2$ #(2)

<const> ::= 1.0 #(0)
         | 0.1 #(1)
         | 10 #(2)
\end{lstlisting}  
\caption{Grammar No bias.}
\label{fig:grammarnoBias}
\end{figure}

In this work, two experiments were carried out to evaluate the performance of the GCGP variants. In the first experiment, the base grammar as shown in Figure \ref{fig:grammarbase}) was used, together with a variant of it where only the division operator was modified. Therefore, this first experiment contemplated two grammars: Gr.Base and Gr.NoDiv.
The second experiment focused on the use of an additional variant of the base grammar, as illustrated in Figure \ref{fig:grammarnoBias}. The purpose of this second experiment was to explore the impact on the generated models. We follow the recommendations about grammar design by Nicolau et al. \cite{Nicolau2018}.

\subsection{Dataset}

We took a subset of NHANES 2017-18 dataset used by Schnur et al \cite{schnur2023information}.  to facilitate accuracy comparisons of the symbolic regression models found in both studies. The data table does not contain neither any individual younger than 18 years old, nor any female pregnant at the moment of the measurements. Individuals with missing demographic, morphometric or objective features were removed from the final dataset. The nine data features used as dependent variables for optimization retain the given names at the original NHANES database and can be seen at Table  \ref{table:dataset}. The model objective was the total body fat measured by dual energy x-ray absorptiometry, named as feature \texttt{DXDTOPF} in the NHANES dataset. We reproduce here the table in \cite{schnur2023information} about the variables and descriptions.
The final dataset contained 2403 unique individuals, with 1158 (48.2\%) males and 1245 (51.8\%) females randomly assigned to training and testing. Moreover, we consider scenarios where we did the training and testing on gender-specific models, so for those, the dataset was split according to the gender column.

\begin{table}[h]
\centering
\begin{tabular}{llccccccc}
\hline
\multicolumn{2}{c}{NHANES} & \multicolumn{7}{c}{Descriptive Statistics} \\
Variable & Description & Mean & Std. & Min. & 25\%tile & 50\%tile & 75\%tile & Max. \\
\hline
SEQN & Anonymous ID Number & - & - & - & - & - & - & - \\
RIAGENDR & Gender (1=`M', 0=`F') & - & - & - & - & - & - & - \\
RIDAGEYR & Age (years) & 38.1 & 12.6 & 18.0 & 27.0 & 38.0 & 49.0 & 59.0 \\
BMXWT & Weight (kg) & 79.7 & 20.4 & 36.2 & 64.9 & 76.9 & 91.9 & 176.5 \\
BMXHT & Height (cm) & 166.6 & 9.3 & 138.3 & 159.4 & 166.5 & 173.8 & 190.2 \\
BMXLEG & Upper Leg Length (cm) & 39.5 & 3.6 & 26.0 & 37.0 & 39.5 & 42.0 & 50.0 \\
BMXARML & Upper Arm Length (cm) & 37.0 & 2.7 & 29.6 & 35.0 & 37.0 & 39.0 & 45.5 \\
BMXARMC & Arm Circumference (cm) & 33.1 & 5.1 & 20.7 & 29.4 & 32.9 & 36.4 & 52.7 \\
BMXWAIST & Waist Circumference (cm) & 96.0 & 16.3 & 56.4 & 83.8 & 94.7 & 106.4 & 154.9 \\
BMXHIP & Hip Circumference (cm) & 104.6 & 12.8 & 77.8 & 95.5 & 102.7 & 111.6 & 168.5 \\
\textbf{DXDTOPF} & \textbf{Total Body Fat \%} & \textbf{33.1} & \textbf{8.6} & \textbf{12.1} & \textbf{27.1} & \textbf{32.9} & \textbf{40.2} & \textbf{56.1} \\
\hline
\end{tabular}
\caption{Variable names and descriptions included in the analysis with descriptive statistics. The target variable is shown in bold.}
\label{table:dataset}
\end{table}

\section{Experimental Results}
\label{sec:exp_results}

\subsection{Experimental setup}

\label{subsec:setup}

{The three variants of GGGP (GE, CFG-GP and DSGE) were configured with specific parameters as detailed in Table \ref{tab:hyperparameters}. The evaluation was performed using the Gr.Base and Gr.NoDiv grammars, with maximum tree depths of 4 and 17, running 10 replicates with populations of 1000 individuals for 1000 generations.
During the training process, the fitness of the best individual was recorded along with the population mean and standard deviation at 10\% intervals of the total generations. The performance of the models was evaluated using the RMSE and R² metrics on both training and test data. The resulting expressions were then simplified using the SymPy library.}

\begin{table}[h!]
    \centering
    \caption{Hyperparameter configuration for GGGP variants as pilot study.}
    \begin{tabular}{|l|c|c|c|}
        \hline
        \textbf{Hyperparameters} & \textbf{GE} & \textbf{DSGE} & \textbf{CFG} \\
        \hline
        \textbf{Runs} & 10 & 10 & 10 \\
        \textbf{Population Size} & 1000 & 1000 & 1000 \\
        \textbf{\# Generations} & 1000 & 1000 & 1000 \\
        \textbf{Probability Crossover} & 0.9 & 0.9 & 0.9 \\
        \textbf{Probability Mutation} & 0.05 & 0.05 & 0.05 \\
        \textbf{Max Tree Depth or Max \# Wraps}& [4, 17] & [4, 17] & [4, 17]\\
        \textbf{Grammars} & [Base, NoDiv] & [Base, NoDiv] & [Base, NoDiv] \\
        \hline
    \end{tabular}
    \label{tab:hyperparameters}
\end{table}

{
We performed simulations with DSGE and CFG methods to study the capacity to obtain best-fitting models and its ability to efficiently explore the solution space.
Now, each GGGP variant was tested using only a modified unbiased grammar, and four values for \emph{maximum tree depth}, either 8, 10, 12 or 14.
Common evolutionary hyperparameters were modified to use 512 individuals through 5000 generations, for 30 runs.
Also, crossover probability was now set to 0.75.
The configurations for this experiment are shown in Table \ref{tab:hyperparameters_tests}.
}

\begin{table}[h!]
    \centering
    \caption{Hyperparameter configuration for 2 GGGP variants exploitation.}
    \begin{tabular}{|l|c|c|}
        \hline
        \textbf{Hyperparameter} & \textbf{DSGE} & \textbf{CFG} \\
        \hline
        \textbf{Runs}  & 30 & 30 \\
        \textbf{Population Size}  & 512 & 512 \\
        \textbf{Generations} & 5000 & 5000 \\
        \textbf{Probability Crossover}  & 0.75 & 0.75 \\
        \textbf{Probability Mutation} & 0.05 & 0.05 \\
        \textbf{Max Tree Depth} & [8, 10, 12, 14] & [8, 10, 12, 14] \\
        \textbf{Grammars} & NoBias & NoBias \\
        \hline
    \end{tabular}
    \label{tab:hyperparameters_tests}
\end{table}

\subsection{Performance Results}
\label{sec:performance}
In this section, we analyze experimental results for the GGGP variants performance on different configurations.
We use RMSE metrics boxplots to visually compare fitness distributions under the various experimental conditions, providing a detailed perspective on the dispersion, central tendency, and stability of the results under those different scenarios.

\begin{figure*}[ht]
\centering
\includegraphics[width=0.85\textwidth]{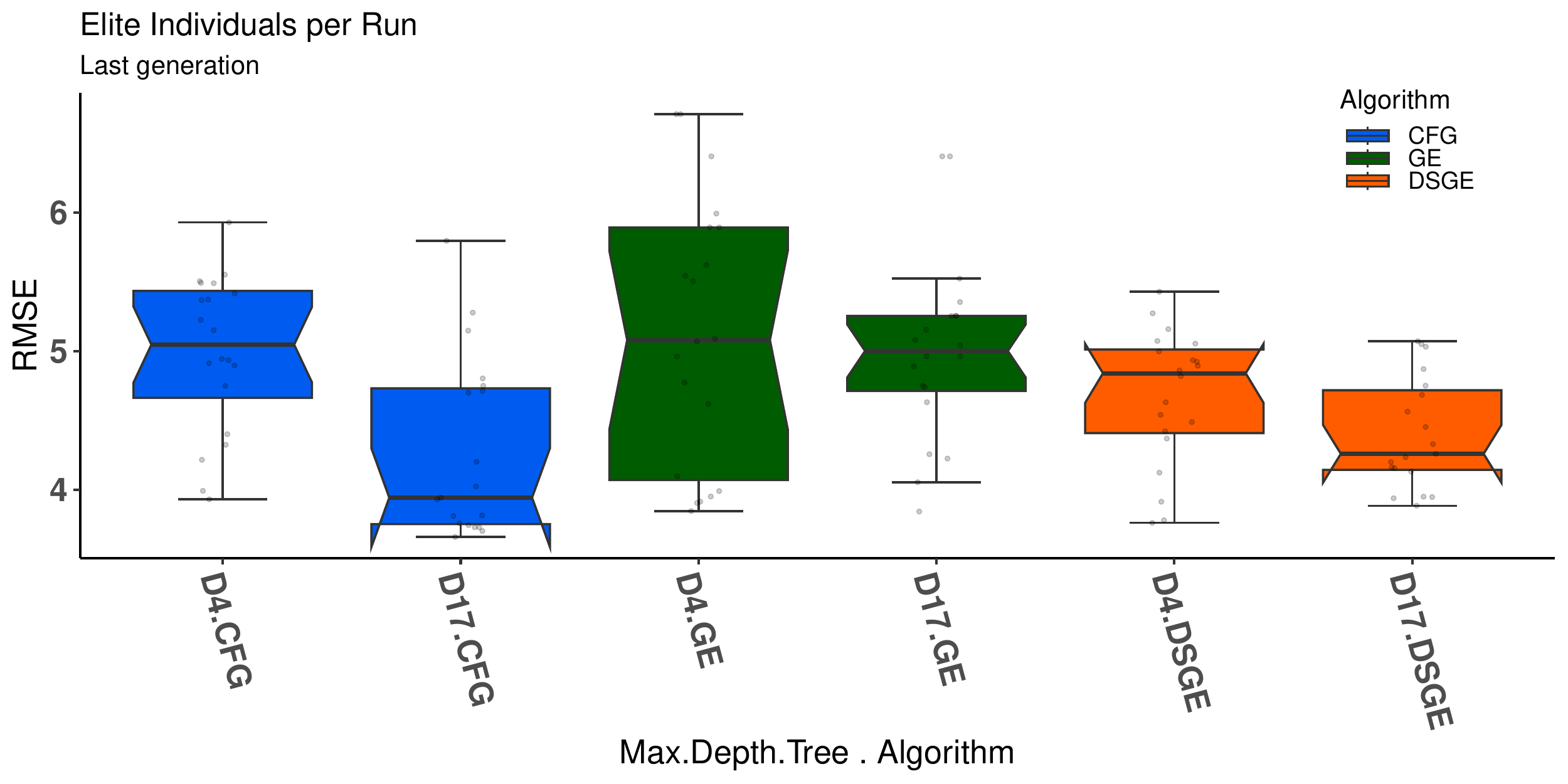}
 \caption{Comparison of the distribution of RMSE of elite solutions at last generation by configuration for depth values of 4 and 17. X-axis legend indicates Max-tree-depth.GGGP-method.}
\label{fig:PopEvol_experiment1}
\end{figure*}

Our first experiment compares analysis of CFG, GE an DSGE, revealing substantial patterns in performance at several maximum tree depths, D4 and D17. 
Results shows, Figure \ref{fig:PopEvol_experiment1}, DSGE consistently surpass the other two GGGP variants, exhibiting lowest RMSE values and lower variability at both depths, followed by GE and CFG respectively.
Tree depth plays a crucial role: D17 tends to produce more accurate results but with higher variability, while D4 shows less dispersion and higher RMSE values.

This evaluation unveils that DSGE represents a more robust option with respect to fitness, particularly when a tradeoff between consistency and precision is required.

\begin{figure*}[ht]
\centering
\includegraphics[width=0.85\textwidth]{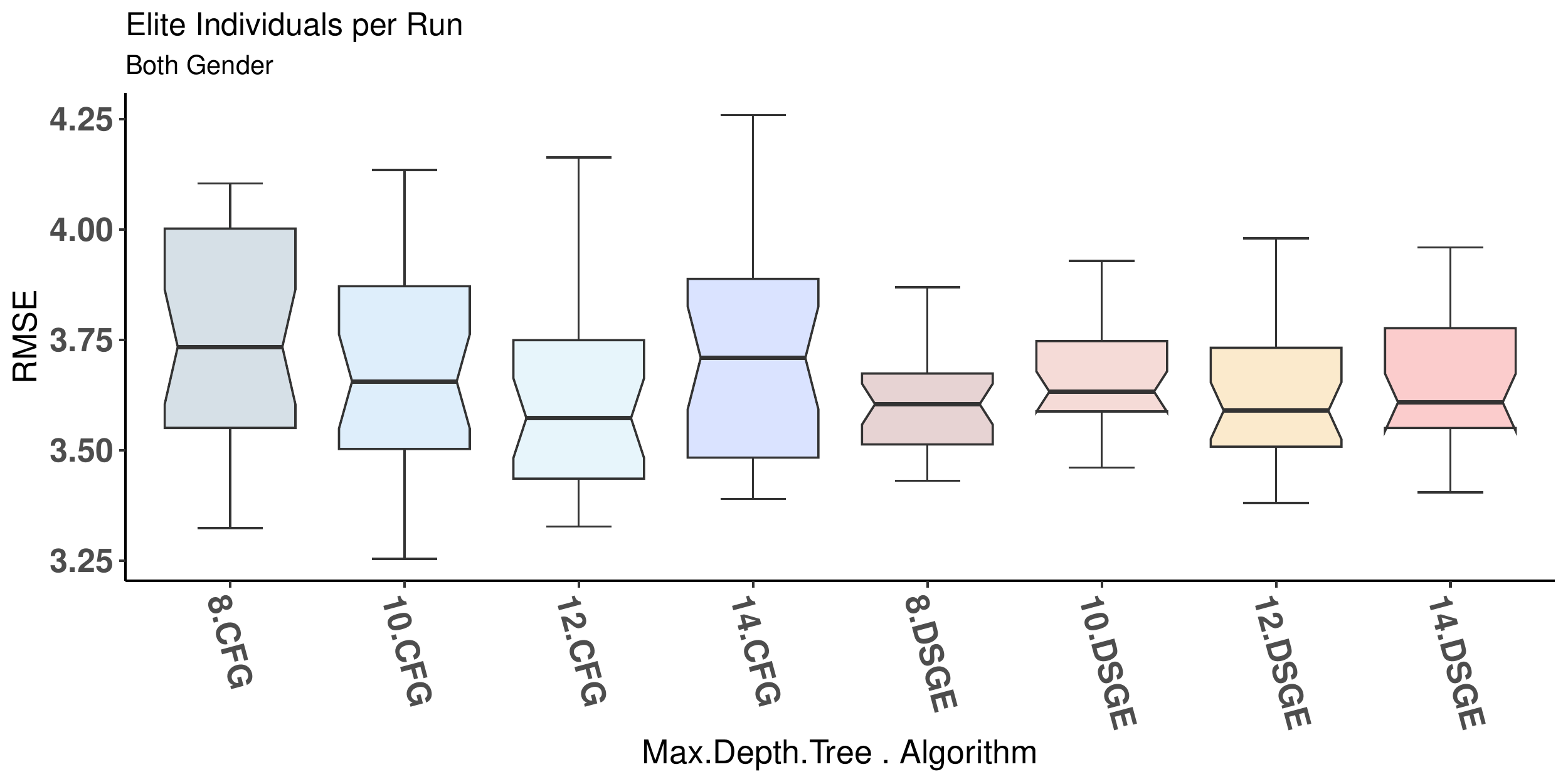}
 \caption{Comparison of the distribution of RMSE of elite solutions at last generation by configuration for depth values of 8, 12 and 12. X-axis legend indicates Max-tree-depth.GGGP-method.}
\label{fig:PopEvol_experiment2}
\end{figure*}

In Figure \ref{fig:PopEvol_experiment2} the performance of the best solutions split by configuration is shown.  Overall, CFG-based solutions were more dispersed for all max. tree depth values than their DSGE counterparts.
Interestingly, minimum fitness values tended to be lower for CFG than DSGE, 

In summary, the results highlight notable differences between the CFG and DSGE methods in terms of performance variability and fitness outcomes across different configurations. CFG-based solutions consistently displayed greater dispersion in fitness values, reflecting a higher variability in performance, regardless of the maximum tree depth. However, this variability also allowed CFG to achieve lower minimum fitness values compared to DSGE, indicating that it could produce some exceptionally high-quality solutions. These findings suggest that while DSGE provides more stable and consistent results, CFG may offer advantages in exploring a broader solution space and achieving superior outcomes in certain scenarios.

Overall, our findings indicate that DSGE is a more reliable method in terms of stability, consistently achieving low RMSE values with minimal variability across all tree depths. While the CFG method is capable of discovering competitive solutions, its higher level of dispersion may be less desirable in applications requiring consistent performance. Thus, DSGE emerges as the preferred choice for generating stable solution sets, whereas CFG, despite its exploratory potential, may exhibit variability that limits its suitability in scenarios where accuracy and consistency are critical.

\subsection{Convergence Analysis}
\label{sec:convergence}
To assess whether the GGGP variants converge toward high-quality solutions, we used the configurations outlined in Table \ref{tab:hyperparameters}.

\begin{figure}[ht]
    \centering
        \includegraphics[width=\linewidth]{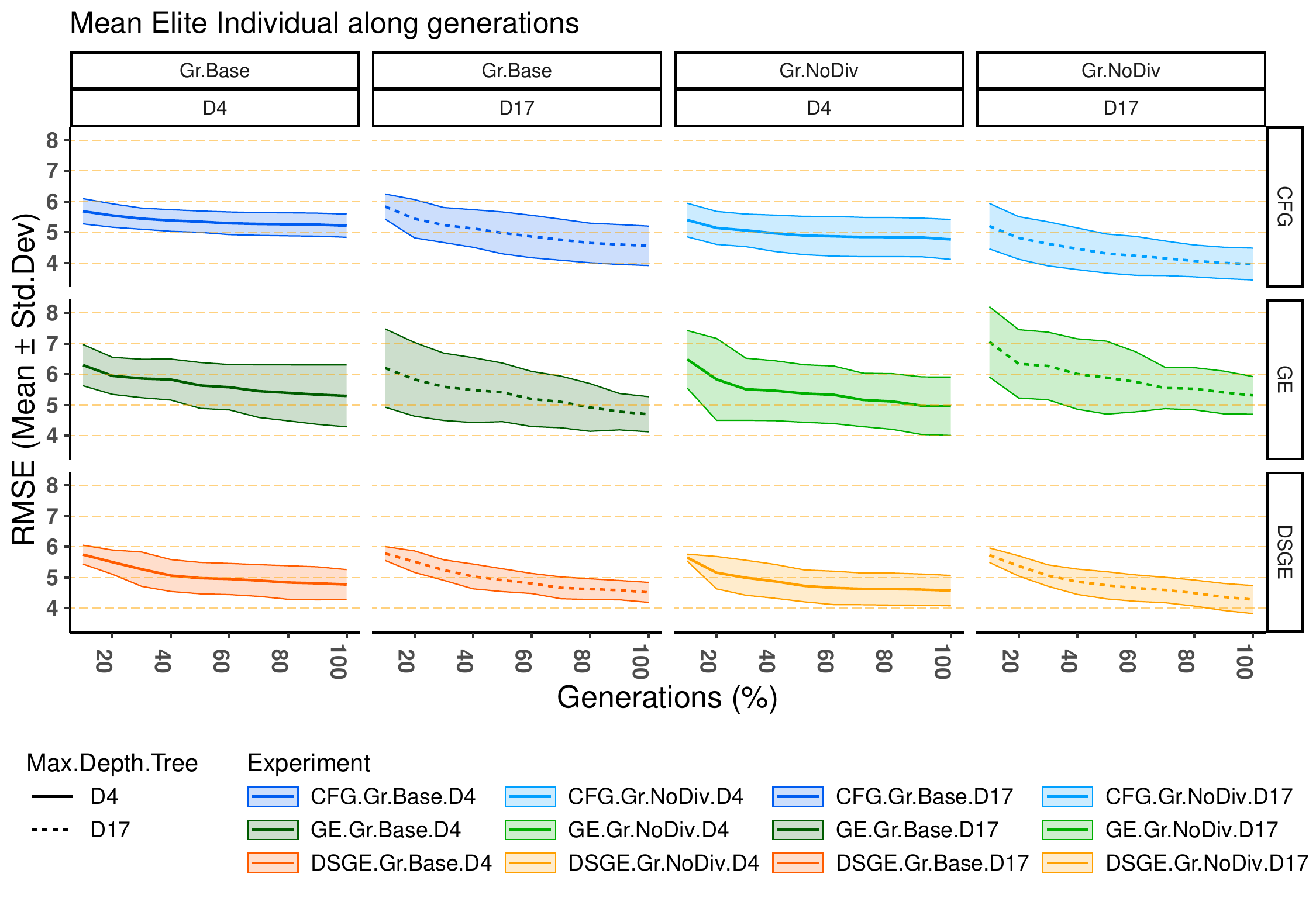} 
        \caption{Time evolution of mean $\pm$  std.dev. of the best solutions pooled by parameter set.}
        \label{fig:Convergencia_experimento1}
\end{figure}

Across all configurations, Figure \ref{fig:Convergencia_experimento1}, the best solutions showed a trend of evolving toward lower fitness values, with configurations set to a maximum tree depth of 17 achieving faster decreases compared to those with a depth of 4. Initial elite fitness values ranged between 5 and 6 in the first generation, eventually decreasing to around 5 for CFG and GE at depth 4, and slightly lower for depth 17. For the DSGE variant, best solutions began at similar fitness values but decreased more rapidly, reaching values near or below 4.

\begin{figure}[ht]
        \centering
        \includegraphics[width=\linewidth]{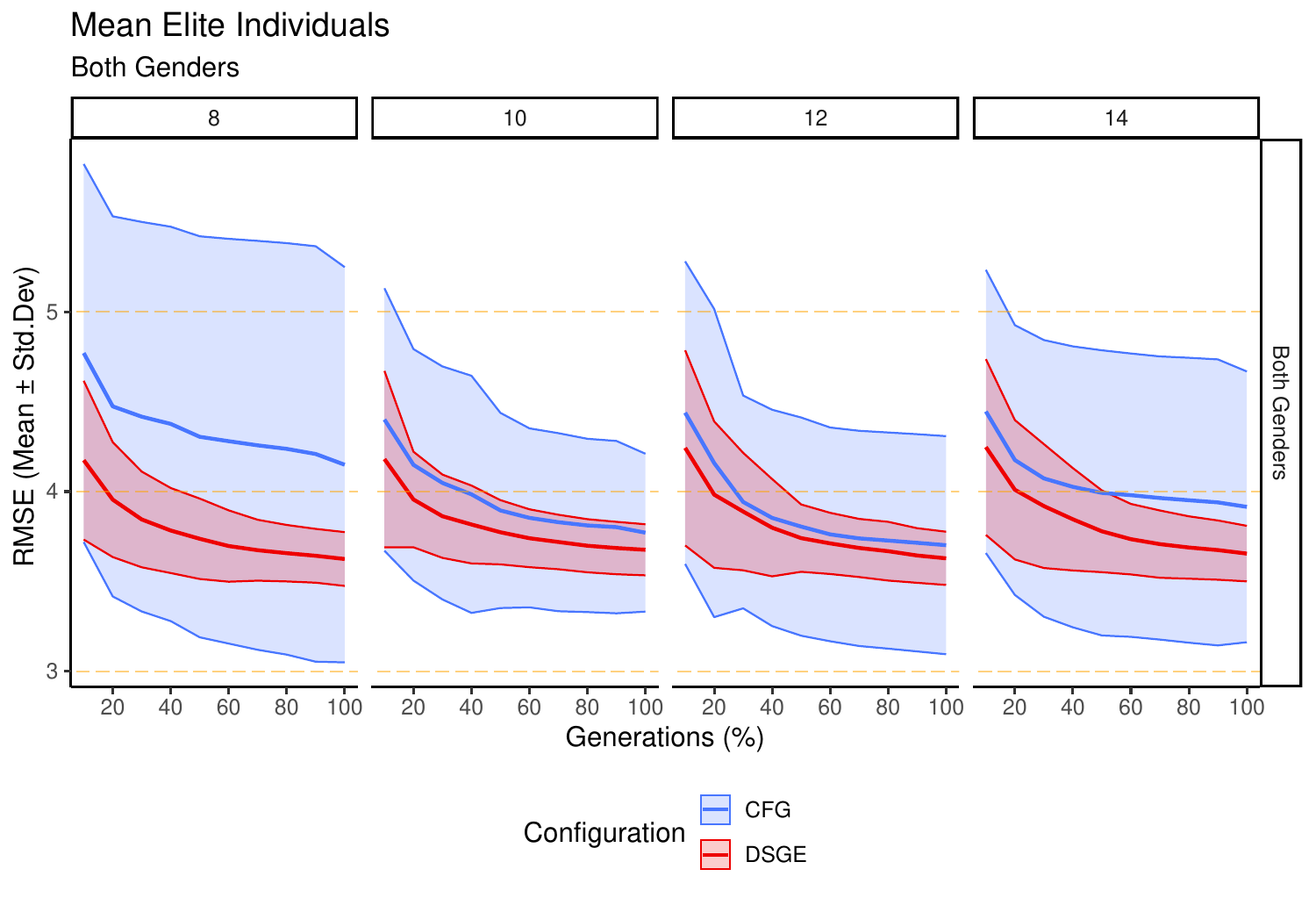} 
        \caption{Re-arrangement of previous figure to facilitate visualization of GGGP algorithm convergence regarding gender split subset and whole dataset .} \label{fig:Results2b}
\end{figure}

Based on the results of the previous experiment, we decided to further analyze the CFG and DSGE variants by adjusting the parameter settings as in Table \ref{tab:hyperparameters_tests}. Figure \ref{fig:Results2b} shows the convergence pattern of both variants under these settings.
Both variants, CFG and DSGE, reduce their mean RMSE, as generations progress, following certain power decay distribution.
CFG-based configurations tends to reach a plateau faster than DSGE, so the 
their average RMSE remain higher.

DSGE stands out for presenting a lower mean RMSE and lower standard deviations in all configurations, indicating a more efficient exploration and exploitation of the search space.
On the other hand, CFG shows a higher population diversity, reflected in higher standard deviations, especially in configurations with greater tree depth.

In conclusion, DSGE is the most suitable option for applications that require stability, due to its lower RMSE and lower variability in solutions. CFG, on the other hand, is beneficial in initial configurations where a more diverse exploration is sought. Greater tree depths allow finding better quality solutions, although at the cost of a longer convergence time, while lower depths offer faster but less accurate results.

The results reveal that a tree depth of 17 consistently outperformed a depth of 4 for both CFG and DSGE. Deeper trees, such as those with depths of 12, 14, and 17, produced lower RMSE values but required more generations to converge. In contrast, shallower trees, with depths of 8 or 10, converged more quickly but yielded solutions with higher RMSE values.

\subsection{Best Models}
Each configuration yields 30 candidate models, corresponding to the lowest RMSE model when fitting the training dataset using the given expression model.
Now, we study models performance of the best solution per configuration based in the lowest RMSE value.
We apply the model to the test dataset and again to the training one and report the $R^2$ metric. Best solutions are summarized in Table \ref{tab:BestSolutions}.

The lowest training RMSE was obtained for the CFG methods with max. depth tree set as 10, which was 3.25 for the training set and 3.29 for the test set and regarding $R^2$, 0.86 for the training dataset and 0.85 for the test one.
However, the resulting formulae obtained had 31 rational terms up to degree 7.
The formula is hence considered unexplainable due to its complexity.

\begin{longtable}{lllllll}
\caption{Best solution set, per configuration; sorted by training dataset fitness (RMSE)} \label{tab:BestSolutions} \\

\hline
 RMSE   & R\textsuperscript{2}    & Avg. Error & Algorithm & Max Tree Depth & Dataset \\ 
\hline
\endfirsthead

\hline
 RMSE   & R\textsuperscript{2}    & Avg. Error & Algorithm & Max Tree Depth & Dataset \\ 
\hline
\endhead

\hline
\endfoot

\hline \hline
\endlastfoot

 3.2539 & 0.8571 & 2.5713     & CFG       & 10             & train   \\
 3.2921 & 0.8489 & 2.5516     & CFG       & 10             & test    \\
 3.3231 & 0.8508 & 2.6288     & CFG       & 8              & train   \\
 3.3409 & 0.8444 & 2.6032     & CFG       & 8              & test    \\
 3.3266 & 0.8504 & 2.6165     & CFG       & 12             & train   \\
 3.3642 & 0.8422 & 2.6064     & CFG       & 12             & test    \\
 3.3801 & 0.8460 & 2.6772     & DSGE      & 12             & train   \\
 3.4191 & 0.8373 & 2.6698     & DSGE      & 12             & test    \\
 3.3891 & 0.8448 & 2.6774     & CFG       & 14             & train   \\
 3.4176 & 0.8372 & 2.6689     & CFG       & 14             & test    \\
 3.4026 & 0.8447 & 2.6915     & DSGE      & 14             & train   \\
 3.4488 & 0.8348 & 2.7084     & DSGE      & 14             & test    \\
 3.4308 & 0.8411 & 2.7063     & DSGE      & 8              & train   \\
 3.4379 & 0.8354 & 2.7058     & DSGE      & 8              & test    \\
 3.4608 & 0.8381 & 2.7362     & DSGE      & 10             & train   \\
 3.4467 & 0.8348 & 2.7042     & DSGE      & 10             & test    \\

\end{longtable}
However, there was a simpler, more interpretable model with fitness, RMSE, of 3.38 for training and 3.41 for test, and $R^2$ of 0.85 for training and 0.84 for test. The simplified formulae obtained was: 

\begin{multline*}
\text{DXDTOPF} = \\
\frac{\text{31}\cdot\text{BMXHIP}}{\text{100}} + 
\frac{\text{9}\cdot\text{BMXHT}\cdot\text{BMXWAIST}}{\text{100000}} - \\
\frac{\text{1387}\cdot\text{BMXHT}}{(\text{130}\cdot\text{BMXWAIST})} -
\frac{\text{BMXWAIST}\cdot\text{BMXWT}^2\cdot\text{RIAGENDR}}{\text{540000}} + \\
\frac{\text{48}\cdot\text{RIAGENDR}}{\text{5}} +
\frac{\text{BMXHT}\cdot\text{BMXWAIST}}{(\text{BMXARML}\cdot\text{BMXWT})}    
\end{multline*}

This solution correspond to the DSGE method with max.tree depth of 12.

\section{Conclusions}
\label{sec:conclusions}

In this paper we have presented an evaluation of several grammar-guided genetic programming (GGGP) variants for symbolic regression to find an interpretable model to predict body fat percentage from morphometric body variables.

We selected the methods called Context-Free Grammar Genetic Programming (CFG), Grammatical Evolution (GE) and Dynamical Structured Grammatical Evolution (DSGE) and performed a parameter sweep on the maximum tree depth and the effect of few related grammars to evaluate the ability to find such interpretable models.

Our first experimental round, comparing the three GGGP methods against a basic grammar with and without division, and two depths, allowed us to conclude that CFG and DSGE outperformed GE for this purpose.
Our second experimental round aimed to exploit CFG and DSGE methods in a larger set of maximum tree depth and using unbiased grammar.
The resulting set of models were filtered out to retain the best model per configuration based on the RMSE metrics on the training dataset. Once again, the models were assessed according to the $R^2$ metric on training and test sets.

The resulting set of 5 models had $R^2$ near 0.85 on both datasets, comparable with those from \cite{schnur2023information}, where they found machine learning models with $R^2$ of 0.85 for training and 0.87 for test datasets.

Further research will be done to explore evolutionary operators and their parameters to improve their ability to find even better models.
In addition, the NHANES dataset can be modeled using patients’ gender, ethnicity and other features.

\section*{Acknowledgments}
J.I.H acknowledges support by Spanish Government AEI grants PID2021-125549OB-I00 and PDC2022-133429-I0 grant funding by MCIN/AEI /10.13039/501100011033 and European UNion Next Generation EU/ PRTR.

\bibliographystyle{splncs04}

\bibliography{bioinspired}

\end{document}